\documentclass[10pt,journal,compsoc]{IEEEtran}

\ifCLASSOPTIONcompsoc
  \usepackage[nocompress]{cite}
\else
  \usepackage{cite}
\fi

\usepackage[utf8]{inputenc} 
\usepackage[T1]{fontenc}    
\usepackage{hyperref}       
\usepackage{url}            
\usepackage{booktabs}       
\usepackage{amsfonts}       
\usepackage{nicefrac}       
\usepackage{microtype}      
\usepackage{graphicx}
\usepackage{tabularx}
\usepackage{hyperref}
\usepackage{color}
\usepackage{amsmath}
\usepackage{algorithm}  
\usepackage{algpseudocode} 
\usepackage{subfigure}
\usepackage{multirow}
\usepackage{soul}

\ifCLASSINFOpdf
\else
\fi
\begin{document}
%
\title{The Computational Drug Repositioning without Negative Sampling}
%
%
%
%

\author{Xinxing~Yang,~\IEEEmembership{}
        Genke~Yang,~\IEEEmembership{}
        and~Jian~Chu~\IEEEmembership{}

\IEEEcompsocitemizethanks{\IEEEcompsocthanksitem X. Yang, G.Yang and J.Chu are with the Ningbo Artificial Intelligence Institute, Shanghai Jiao Tong University and the Department of Automation, Shanghai Jiaotong University. Email: yangxinxing@sjtu.edu.cn; gkyang@sjtu.edu.cn; chujian@sjtu.edu.cn;}

\thanks{Corresponding Author: Genke~Yang; Email: gkyang@sjtu.edu.cn.}
\thanks{This research was partially funded by the China National R\&D Key Research Program (2020YFB1711204) (2019YFB1705702).}}

%
%

\markboth{Journal of \LaTeX\ Class Files,~Vol.~14, No.~8, August~2015}%
{Shell \MakeLowercase{\textit{et al.}}: Bare Demo of IEEEtran.cls for Computer Society Journals}
%



\IEEEtitleabstractindextext{%
\begin{abstract}
Computational drug repositioning technology is an effective tool to accelerate drug development. Although this technique has been widely used and successful in recent decades, many existing models still suffer from multiple drawbacks such as the massive number of unvalidated drug-disease associations and the inner product. The limitations of these works are mainly due to the following two reasons: firstly, previous works used negative sampling techniques to treat unvalidated drug-disease associations as negative samples, which is invalid in real-world settings; secondly, the inner product cannot fully take into account the feature information contained in the latent factor of drug and disease. In this paper, we propose a novel PUON framework for addressing the above deficiencies, which models the risk estimator of computational drug repositioning only using validated (\textbf{P}ositive) and unvalidated (\textbf{U}nlabelled) drug-disease associations without employing negative sampling techniques. The PUON also proposed an \textbf{O}uter \textbf{N}eighborhood-based classifier for modeling the cross-feature information of the latent facotor. For a comprehensive comparison, we considered 8 popular baselines. Extensive experiments in four real-world datasets showed that PUON model achieved the best performance based on 6 evaluation metrics. \par
\end{abstract}

}

\maketitle

\IEEEdisplaynontitleabstractindextext

%
\IEEEpeerreviewmaketitle

\IEEEraisesectionheading{\section{Introduction}\label{sec:introduction}}

\IEEEPARstart{W}{ith} the rapid development of genomics and bioinformatics in the last half-century, traditional innovative drug development technologies for specific diseases are becoming more and more sophisticated \cite{1,2}. However, innovative drugs need to go through time-consuming, costly, and uncontrollable steps such as compound testing, clinical trials and in vivo experiments before they can be successfully marketed \cite{3,4}. As a result, the development cycle for traditional innovative drugs takes 10-15 years and costs approximately 0.8-1.5 billion \cite{5}. Despite the huge investment of time and money by pharmaceutical companies, the number of drugs approved by the U.S. Food and Drug Administration (FDA) has continued to decline since the 1990s \cite{6,7}. It is worth noting that to date, there are 15,000 real diseases known to mankind in the world, yet only one-third of them can be treated. Hence, there is an urgent need for a shift in the way drugs are developed to better address the diseases that plague us \cite{8,9}. \par

The purpose of computational drug repositioning techniques is to uncover new uses for marketed drugs, i.e., new indications \cite{10}. The principle is that a marketed drug can react with multiple unknown targets and thus will act as a treatment for other unknown diseases. Computational drug repositioning techniques learn the laws behind the therapeutic relationships from validated drug-disease associations, and thus new therapeutic relationships can be inferred. Generally speaking, computational drug repositioning models can be divided into two categories, one is graph-based approach \cite{n1,n2} and the other is latent factor-based approach \cite{m1,m2}. \par

The graph-based approach mainly involves random walk on the drug-disease association network to calculate the probability of association between drug and disease node. For example, Luo et al. \cite{an3} proposed a graph-based method, MBiRW, which constructs a heterogeneous network with multiple nodes and multiple relationships based on drug-disease association information, drug-drug similarity information and disease-disease similarity information. Where the nodes are drugs and diseases and the relationships are treatment relationships and similarity relationships. Subsequently, a bipartite random walk algorithm was used to train on the heterogeneous graph for several iterations to calculate the probability of the existence of treatment relationships between drugs and diseases. Chen et al. \cite{an4} proposed another graph-based method iDrug. They believe that there are many common drugs in the drug-target association network and the drug-disease association network. The node representations of these drugs in the two networks should be close enough. This restriction can make the drug node representation more robust and in line with common sense. Therefore, they built a drug-target-disease heterogeneous network by sharing drugs. Through transfer learning, they can mine new drug-target associations and drug-disease associations on this heterogeneous network. In contrast to previous work, which focuses on drug structure to discover potential indications, Yang et al. \cite{an5} argue that this approach can limit the discovery of new functions, so they embed chemical-genomics and pharmaco-genomics in drug-target networks for the first time to help find potential new drug-target associations, based on the hypothesis that "drug-target associations exist mainly due to drug substructures and protein domains". \par

The latent factor-based models mainly use the latent factor of drugs and diseases to calculate the degree of similarity between them. The existing work mainly explores how to mine the effective latent factor. The generalized matrix factorization (GMF) proposed by He et al. \cite{xn3} is a classic model in the field of recommender systems. The current computational drug repositioning problem can be considered from the perspective of the recommendation system. The GMF model uses existing drug-disease associations to learn the latent factors of drugs and diseases. Subsequently, GMF did not use the inner product operation to infer the predicted value, but instead multiplied the elements of the two latent facto and input them into the neural network. This kind of operation mathematically proved to be an adaptive version of the inner product operation, which has better results and adjustment ability. However, GMF did not consider the domain information about the drug repositioning scenario, which caused its performance to be greatly affected by the sparsity. Based on the GMF model, Yang et al. \cite{xn4} proposed the Additional Neural Matrix Factorization model (ANMF) for finding new uses of existing drugs. The ANMF model embeds the respective similarity information of drugs and diseases into the latent factor space with a large sparseness. The embedding of this dense information makes the latent factor of drugs and diseases avoid the cold start problem. And the use of negative sampling technology improves the effect of the ANMF model. In addition, Yang et al. \cite{xn5} proposed a Hybrid Attentional Memory Network (HAMN) model for the drug repositioning problem. HAMN uses the attention network to cleverly combine the common advantages of the latent factor-based model and the graph-based model so that the predicted value can take into account both global and local connection information. \par

As mentioned in the above approach, there is a default assumption in previous work. That is, in order to populate the training set, unvalidated drug-disease associations are randomly marked as negative samples by negative sampling techniques, which is invalid in real-world settings \cite{11,12,13}. This assumption is not necessarily true for the following reason: unvalidated associations between a drug and a disease simply lack clinical validation and do not necessarily mean that the drug cannot treat the disease. In other words, these unvalidated drug-disease associations are neither positive nor negative samples. If all unvalidated drug-disease associations were labeled as negative samples, this would result in some potentially therapeutic drug-disease associations being mislabeled, leading to the generation of false samples in the training set and the consequent degradation of model prediction performance. \par

Furthermore, the matrix factorization model (latent factor-based) \cite{14}, which is currently the cornerstone model in the field of computational drug repositioning, has shortcomings in its ability to capture more complex drug-disease associations. The reason behind this is that the inner product operation only considers the cumulative information of each dimension of the latent factor in turn and does not model the cross information between the dimensions. \par

Therefore, in order to solve the deficiencies in the existing model, in this work, we propose a novel framework called PUON, in which only the verified drug-disease association (\textbf{P}ositive) and unverified drug-disease association (\textbf{U}nlabelled) are used to train the outer product-based classifier (\textbf{O}uter and \textbf{N}eighborhood) to find the new uses of marketed drugs. The highlight of PUON is its assumption that the set of unvalidated drug-disease associations consists of a certain proportion of positive samples (drug-disease associations with a therapeutic relationship) and a certain proportion of negative samples (drug-disease associations without a therapeutic relationship), i.e., the set of unvalidated drug-disease associations is a weighted sum of positive and negative sample data.  \par

Based on the above assumption, PUON derives a risk estimator for the computational drug repositioning domain without using negative samples. An optimal classifier can be obtained for mining potential drug-disease associations by minimizing the risk estimator. Unlike the classifier based on the matrix factorization model, which uses inner product operation to infer the predicted values, the PUON proposed an Outer Neighborhood-based classifier for modeling the cross-feature information of latent facotors of drugs and diseases to make it more expressive. The Outer Neighborhood-based classifier consists of two parts: outer product and neighborhood collaboration. The derivation in Section 2.3 can be derived that the outer product contains more degrees of freedom compared to the inner product. Besides, the neighborhood collaboration in the classifier draws on the key assumption in drug development that "similar drugs usually treat similar diseases". It believes that similar drugs should have similar treatment probabilities for the same disease. Therefore, for the predicted value between the target drug and the target disease, the classifier additionally considers the score of the most similar neighbor of the target drug, and the collaboration of the above-mentioned neighborhood information essentially enhances feature expressiveness. The main contributions of this work are as follows:\par

\begin{enumerate}
	\item We novelly propose a risk estimator framework for the field of computational drug repositioning that allows the prediction of potential drug-disease associations without the need for negative sampling techniques.
	\item We propose a novel outer neighborhood-based classifier that models the cross-feature information between dimensions of latent factor and enhances feature expressiveness by collaborating with neighborhood information.
	\item We conducted extensive experiments to validate the effectiveness of the above improvement points using four real-world datasets.
\end{enumerate}

\begin{table}[h]
	\caption{The statistics of four real-world datasets}
	\centering
	\setlength{\tabcolsep}{2.5mm}
	\begin{tabular}{@{}ccccc@{}}
		\toprule
		Datasets  & Drugs & Diseases & Validated Associations & Sparsity \\ \midrule
		Gottlieb  & 593   & 313      & 1933                   & 98.95\%   \\
		Cdataset  & 663   & 409      & 2532                   & 99.06\%   \\
		DNdataset & 1490  & 4516     & 1008                   & 99.98\%   \\
		Ldataset  & 598   & 269      & 18416                  & 88.55\%   \\ \bottomrule
	\end{tabular}
\end{table}

\begin{figure*}[th]
	\centering
	\includegraphics[scale=0.5]{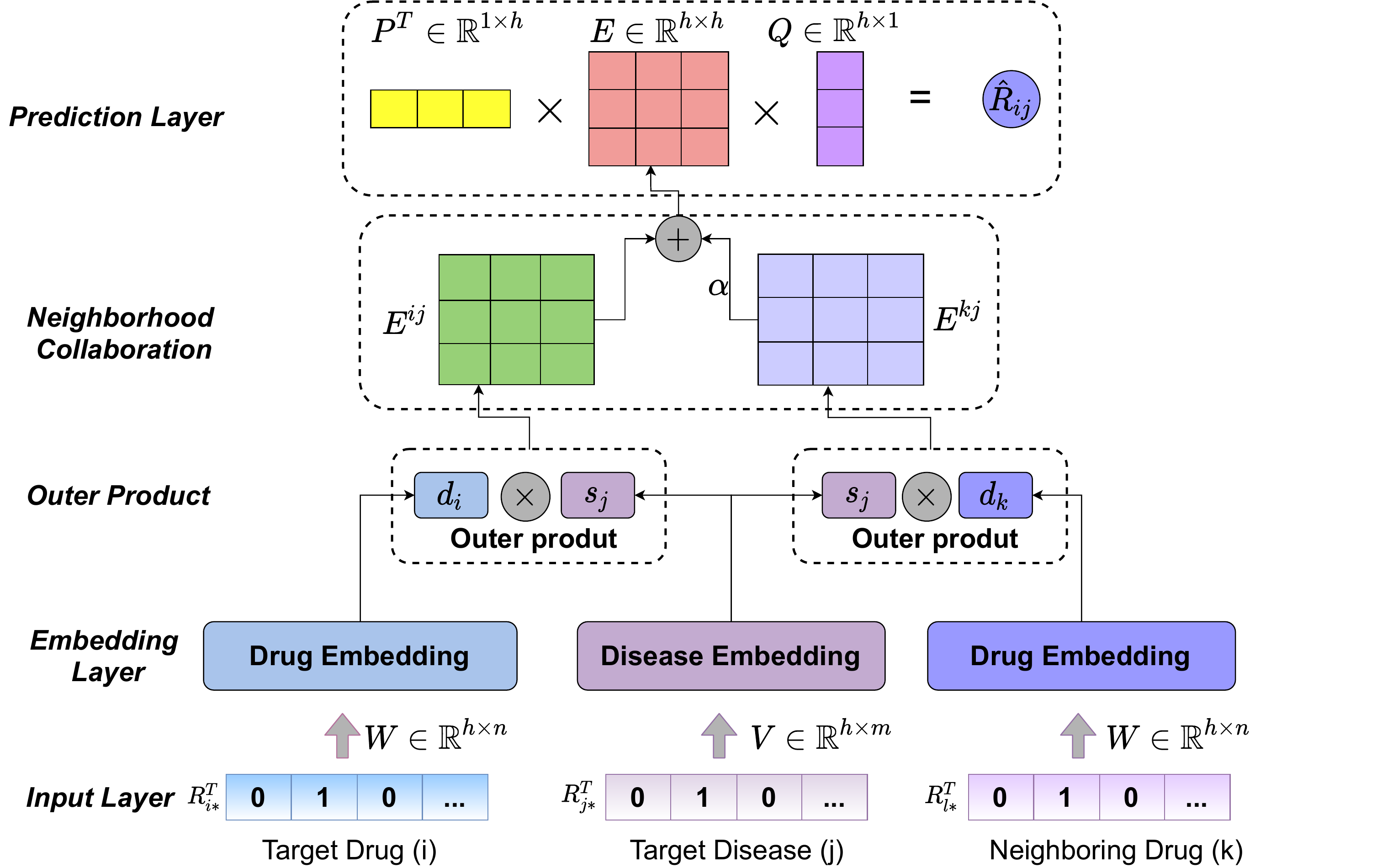}
	\caption{The framework of classifier $C$ in PUON.}
	\label{}
\end{figure*} \par

The rest of this work is organized as follows. The section 2 describes the implementation details of the PUON framework. The section 3 describes the relevant experimental results. The section 4 summarizes the content of this work and the description of future work. \par

\section{Materials and Methods}

In this section, we first introduce the dataset used for the experiments in Section 2.1; The risk estimator of PUON model is introduced in Section 2.2; The classifier based on the outer neighborhood in PUON model is introduced in Section 2.3. \par

\subsection{Datasets}

Four benchmark datasets, Gottlieb dataset \cite{15}, Cdataset \cite{an3}, DNdataset \cite{an3} and Ldataset \cite{add5}, were used in this work. These four datasets vary in the number of drugs, number of diseases, and number of validated associations. The four datasets both are available for download from public websites \cite{an3}. The Gottlieb dataset consists of 593 drugs, 313 diseases, and 1933 validated (drug, disease) association tuples. There is a drug similarity matrix with a shape of (593, 593) and a disease similarity matrix with a shape of (313, 313). The values of the similarity matrix represent the degree of similarity between drugs or between diseases. The Cdataset consists of 663 drugs, 409 diseases, and 2532 validated (drug, disease) association tuples. And Cdataset also contains a drug-to-drug similarity matrix and a disease-to-disease similarity matrix. The DNdataset consists of 1490 drugs, 4516 diseases, and 1008 validated (drug, disease) association tuples. The Ldataset consists of 598 drugs, 269 diseases, and 18416 validated (drug, disease) association tuples.\par

The relevant information about the above-mentioned drugs and diseases comes from the DrugBank dataset \cite{drugbank} and the Online Mendelian Inheritance in Man (OMIM) database \cite{omim}, respectively. The similarity matrix of drugs or diseases is provided by the work of Luo et al \cite{an3,cdk,smile,add5}. In addition, The similarity matrix between drugs, $DrugSim$, which is calculated from the SMILES chemical structures of the drugs. The similarity matrix between diseases, $DisSim$, which is calculated from the medical description information between them. All the above data can be downloaded from the public website. The data statistics of the above four datasets are shown in Table 1.\par

\begin{figure*}[h]
	\centering
	\includegraphics[scale=0.6]{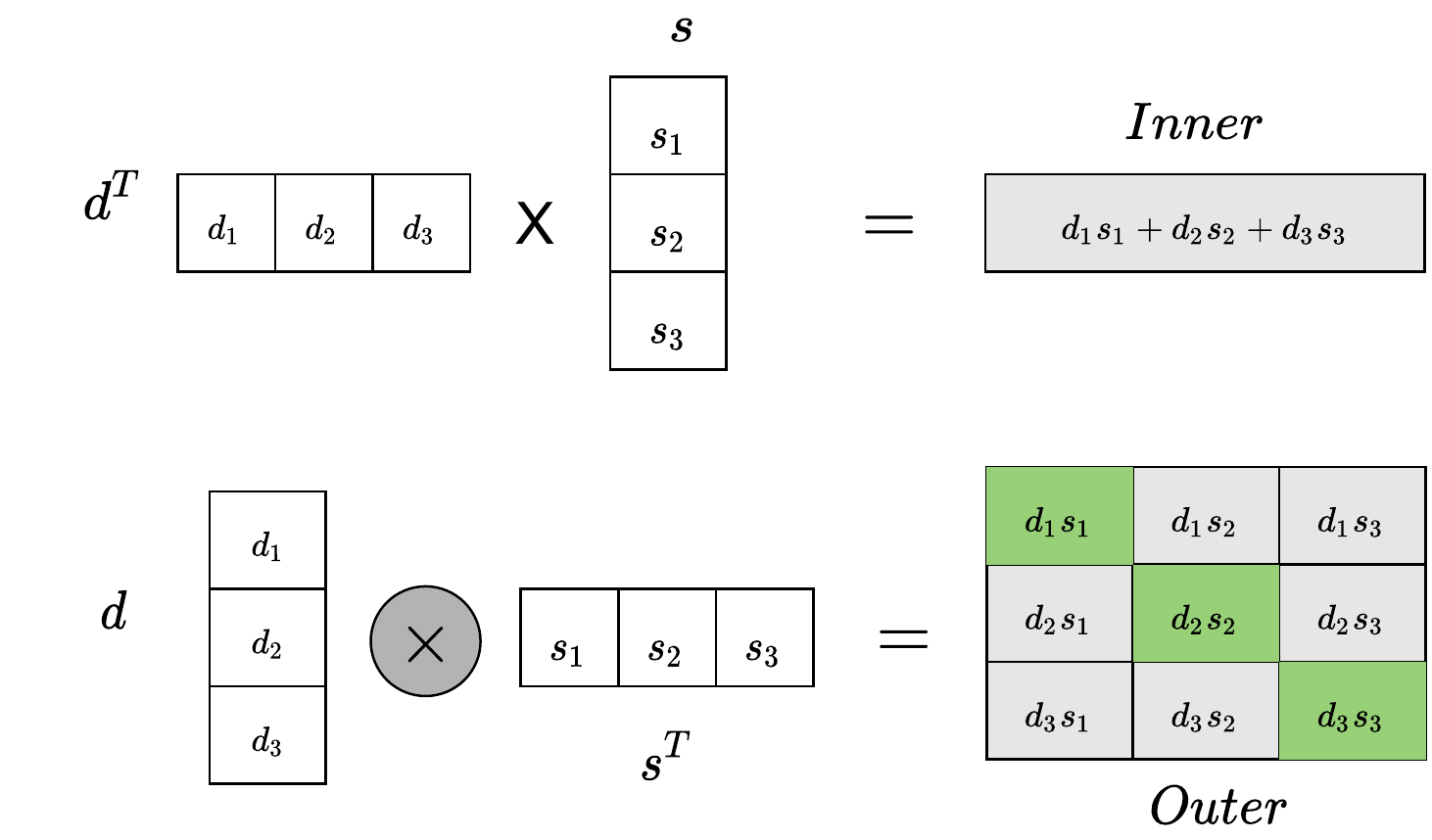}
	\caption{The example of comparing inner product and outer product.}
	\label{}
\end{figure*} \par

\subsection{The Risk Estimator for Computational Drug Repositioning without Negative Samples}

In the area of computational drug repositioning, the goal of PUON is to learn a classifier $C$, (either a matrix factorization model or a graph model) to map each drug-disease association tuple $(d,s)$ to a scalar value $label$ that represents the probability that drug $d$ can treat disease $s$. $d$ is the initials of the drug and $s$ is the initials of the sickness. And $label = C(d,s), label \in{\{0,1\}}$. In order to overcome the drawbacks of the negative sampling technique, we design the risk estimator without negative samples for the computational drug repositioning as shown below. \par

We first let $p_{data}(d,s)$ be the underlying joint distribution of the drug-disease association tuple, and $\pi_{p} = p(label = 1)$ be the positive class prior of the drug-disease association tuple. Therefore, according to the law of the total probability, $p_{data}(d,s)$ can be rewritten in the form as shown in Equation (1). \par

\begin{align}
	p_{data}(d,s) &= \pi_{p} p_{p}(d,s) +(1-\pi_{p})p_n(d,s) \\ \notag
	&= p(label = 1) p_{data}(d,s| label =1) \\ \notag
	&+ (1-p(label = 1))p_{data}(d,s| label =0)
\end{align}

The above equation assumes that the positive drug-disease association tuple (i.e., there is a treatment relationship) is sampled from the positive marginal distribution $ p_{p}(d,s) = p_{data}(d,s| label =1)$. The negative drug-disease association tuple (i.e., there is no treatment relationship) is sampled from the negative marginal distribution $ p_{n}(d,s) = p_{data}(d,s| label =0)$. \par

In general, the risk estimator of PUON is $R(C) = E_{p_{data(d,s)}} \big[L\big(C(d,s),label^t\big)\big] $ when based on a specific loss function $L(y,\hat{y})$. This risk estimator in essence computes an error between the result of the classifier $C$ and the true result $label^t$. Therefore, in order to learn an optimal classifier $C$, we need to minimize the risk estimator. Based on the underlying joint distribution of the drug-disease association tuple obtained from the Equation (1), $R(C)$ can be rewritten as shown in Equation (2) below. \par

\begin{align}
	R(C) &= \pi_{p} R^{+}_{p}(C) +(1-\pi_{p}) R^{-}_{n}(C) \\ \notag
	&=\pi_{p} E_{p_{p(d,s)}} \big[L\big(C(d,s),1\big)\big] \\ \notag
	&= (1-\pi_{p}) E_{p_{n(d,s)}} \big[L\big(C(d,s),0\big)\big] 
\end{align}

The formula is mainly divided into two parts, one is $\pi_{p} R^{+}_{p}(C) = \pi_{p} E_{p_{p(d,s)}} \big[L\big(C(d,s),1\big)\big]$, which is  risk of the classifier $C$ on the set of drug-disease associations with treatment relationship. The other is $(1-\pi_{p}) R^{-}_{n}(C) = (1-\pi_{p}) E_{p_{n(d,s)}} \big[L\big(C(d,s),0\big)\big] $, which is the risk of the classifier $C$ on the set of drug-disease associations without treatment relationship. By the above equation we can find that the training data for $ R^{+}_{p}(C)$ can be drawn from the set of known drug-disease associations. However, the negative samples required in $ R^{-}_{n}(C)$, we do not have access source. \par

In previous computational drug repositioning studies, negative sampling techniques have been used to mark unvalidated drug-disease associations as negative samples. However, an unvalidated association between a drug and a disease is simply a lack of clinical validation and does not necessarily mean that the drug cannot treat the disease. Simply labeling unvalidated drug-disease associations as negative samples can result in a model with unstable and poor generalization performance. \par

Therefore, PUON assumes that the set of unvalidated drug-disease associations consists of a certain proportion of positive samples (drug-disease associations with a treatment relationship) and a certain proportion of negative samples (drug-disease associations without a treatment relationship), i.e., it is a weighted set of positive and negative sample data. Therefore, according to this assumption, we can obtain the marginal distribution of the set of unvalidated drug-disease associations, $ p_{u}(d,s) = \pi_{p} p_{p}(d,s) + (1-\pi_{p}) p_n(d,s) $. Where $u$ is the set of unvalidated drug-disease associations. Then the marginal probability of negative sample can be written as, $ (1-\pi_{p}) p_n(d,s) = p_{u}(d,s) - \pi_{p} p_{p}(d,s) $. Then $ R^{-}_{n}(C)$ can be rewritten as shown in the following equation. \par

\begin{align}
	(1-\pi_{p}) R^{-}_{n}(C)  = R^{-}_{u}(C) - \pi_{p}  R^{-}_{p}(C) 
\end{align}

Equation (3) contains two parts, one is the risk of unlabeled samples being labeled as negative samples, and the other is the risk of positive samples being labeled as negative samples. therefore, by the derivation of the above equation, the final risk estimator of PUON can be rewritten as follows. \par

\begin{align}
	R(C) &= \pi_{p} R^{+}_{p}(C) - \pi_{p} R^{-}_{p}(C) +R^{-}_{u}(C) \\ \notag
	&=\pi_{p} E_{p_{p(d,s)}} \big[L\big(C(d,s),1\big)\big] \\ \notag
	&- \pi_{p} E_{p_{p(d,s)}} \big[L\big(C(d,s),0\big)\big] \\ \notag
	&+ E_{p_{u(d,s)}} \big[L\big(C(d,s),0\big)\big] 
\end{align}

In this work, the loss function $L$ in Equation (4) uses the cross-entropy loss, i.e., $L(y^{p},y^{l}) = -\big[ y^{l} \log y^{p} + (1-y^{l}) \log (1-y^{p}) \big]$, where $y^{p}$ is the predicted value of the model and $y^{l}$ is the true label value. By replacing the cross-entropy loss $L$ into Equation (4), the final optimization objective $R(C)$ can be rewritten as shown in Equation (5).  \par

\begin{align}
R(C) &= - \big[ \frac{1}{N_p} \sum_{(d,s) \in \Omega^p}^{N_p} \big( \pi_{p}\log C(d,s) - \pi_{p} log(1-C(d,s)) \big) \\ \notag
&+ \frac{1}{N_u} \sum_{(d,s) \in \Omega^u}^{N_u} \log(1-C(d,s)) \big] 
\end{align}

Where $\Omega^p$ denotes the set of positive samples used for training, which are sampled from validated drug-disease associations. And $\Omega^u$ denotes the set of unlabeled samples used for training, which are sampled from unvalidated drug-disease associations. $N_p$ is the number of samples in $\Omega^p$ and $N_u$ is the number of samples in $\Omega^u$. By minimizing the above Equation (5), we can then obtain the optimal classifier $C$ without using negative samples. In the subsequent experimental sections, we find that our proposed risk estimator is more suitable for the computational drug repositioning problem than the negative sampling techniques, with better performance and avoidance of overfitting situations.  \par

\subsection{Outer Neighborhood-based Classifier}

The classifiers $C$ in the existing works are based on matrix factorization models. The matrix factorization model uses parameter vector $d_i \in \mathbb{R}^{h \times 1}$ with the dimensionality of $h$ represents the latent factor of drug $i$ and $s_j \in \mathbb{R}^{h \times 1}$ represents the latent factor of disease $j$. Subsequently, $d_i$ and $s_j$ are subjected to the inner product operation to infer the probability that drug $i$ can cure disease $j$. However, the drawback of the inner product operation is that it only considers the cumulative information of each dimension of the latent factor and does not model the cross information between each dimension. \par

Therefore, unlike the classifier based on the matrix factorization model, which uses inner product operation to infer the predicted values, the PUON proposed an outer neighborhood-based classifier for modeling the cross-feature information of latent facotors (drugs and diseases) to make it more expressive. As shown in Figure 1, the outer neighborhood-based classifier consists of five parts: input layer, embedding layer, outer product, neighborhood collaboration and prediction layer. \par

\textbf{Input and Embedding Layer.} Firstly, the PUON mines the latent factor of the target drug $i$, the neighboring drug $k$, and the target disease $j$ using the therapeutic behavior of the drug and the disease as input. The operational formulas are shown in Equation (6) and Equation (7).

\begin{align}
	\big\{ d_{i}, d_{k} \} &= f(W \big\{ R_{i*}^T, R_{k*}^T \}+ b_1)  \\ \notag
	k &\in \max\limits_k DrugSim_{i,k}
\end{align}

\begin{equation}
	s_{j} = f(V R_{j*}^T + b_2)
\end{equation}

Where $d_i \in \mathbb{R}^{h\times 1}$ and $d_k \in \mathbb{R}^{h\times 1} $ with the dimensionality of $h$ are the latent factor of the target drug $i$ and neighboring drug $k$, respectively, and neighboring drug k is the drug with the largest similarity to drug i in the similarity matrix $DrugSim$, where the DrugSim matrix represents the drug-drug similarity matrix, and its similarity is calculated from the SMILES strings between drugs. $s_j \in \mathbb{R}^{1\times h} $ with the dimensionality of $h$ is the latent factor of the target disease $j$. In addition, $R\in \mathbb{R}^{m\times n}$ is the drug-disease association matrix, $ R_{i*}\in \mathbb{R}^{1\times n} = [R_{i1}, R_{i2},..., R_{in}]$, $ R_{k*}\in \mathbb{R}^{1\times n} = [R_{k1}, R_{k2},..., R_{kn}]$ and $ R^T_{j*}\in \mathbb{R}^{1\times m} = [R^T_{j1}, R^T_{j2},..., R^T_{jm}]$, $m$ represents the number of drugs and $n$ represents the number of diseases. $f$ is the activation function, $W \in \mathbb{R}^{h\times n}$ and $V \in \mathbb{R}^{h\times m}$ are the weight parameters in the neural network, and $b_1 \in \mathbb{R}^{h\times 1}$ and $b_2 \in \mathbb{R}^{h\times 1}$ are the bias parameter. \par

\textbf{Outer Product.} Then PUON performs an outer product operation on the target drug $i$ and the target disease $j$ using Equation (8) to get the interaction matrix $E^{ij} \in \mathbb{R}^{h\times h}$. Figure 2 shows that the outer product contains more feature information compared to the inner product. \par

\begin{equation}
	E^{ij} = d_i \otimes s_j = d_i s_j^T
\end{equation}

The $\otimes $ is the outer product operation. As shown in the example in Figure 2, the sum of the values of the main diagonal of the interaction matrix $E^{ij}$ obtained by the outer product is equivalent to the result of the inner product operation, so it can be concluded that the outer product operation possesses more feature information.  \par

\textbf{Neighborhood Collaboration.} Previous work has assumed that similar drugs should have similar treatment probabilities for the same disease. Therefore, as shown in Equation (9), for the predicted values between the target drug $i$ and the target disease $j$, PUON additionally use the interaction matrix $E^{kj} \in \mathbb{R}^{h\times h}$ of neighboring drugs $k$ and diseases $j$ for information collaboration. $\alpha$ is a moderating factor to adjust the weight of $E^{kj}$. The collaboration of the above neighborhood information make the PUON model take into account more feature information. Compared with the inner product, the interaction matrix $E$ contains more features that are used to infer the predicted probability of whether the drug can treat the disease. \par 

\begin{align}
E^{kj} &= d_k \otimes s_j \\ \notag
E &= E^{ij} + \alpha E^{kj}
\end{align}

\textbf{Prediction Layer.} Finally, as shown in Equation (10), PUON model uses two memory matrices $P \in \mathbb{R}^{h\times 1}$ and $Q \in \mathbb{R}^{h\times 1}$ to calculate the predicted treatment probability of the target drug and target disease, and they can automatically learn the appropriate weights for the features in $E$.  In addition to this, we can convert the interaction matrix $ E $ into a vector with a row value of 1. It is then fed into a single-layer neural network to derive the predicted probability that the target drug $i$ can treat the target disease $j$.\par

\begin{equation}
	\hat{R}_{ij} \leftarrow F(P^T E Q+b)
\end{equation}

Where $\hat{R}_{ij}$ is the predicted treatment probability. The pseudo-code for PUON is shown in Algorithm 1.\par

\subsection{Optimization}

In order to learn the parameters in the classifier $C$, we need to minimize the risk estimator $R(C)$. And we need to obtain relevant samples by sampling technique in order to facilitate the optimization of $R(C)$. The data sampling strategy of PUON as shown in Algorithm 1.\par
The number of positive samples in the training set is denoted by $N_p$. The number of unlabeled samples in the training set is $N_u$. First, we need to sample $N_p$ (drug, disease) tuples from the set of validated drug-disease associations with label 1 for training the first part $\pi_{p} R^{+}_{p}(C) $ of $R(C)$. Subsequently, another $N_p$ (drug, disease) tuples from the set of validated drug-disease associations need to be sampled with label 0 for training the second part $- \pi_{p} R^{-}_{p}(C)$ of $R(C)$. Subsequently, $N_u$ (drug, disease) tuples from the set of unvalidated drug-disease associations need to be resampled with label 0 for training the third part of $R(C)$, $ R^{-}_{u}(C) $. \par

By iterating the above steps, we can minimize $R(C)$ and obtain the trained classifier $C$. \par

\begin{algorithm}
	\caption{PUON.} 
	\textbf{Input-1:} Drug-Disease association matrix, $R$\\
	\textbf{Input-2:} Drug-Drug similarity matrix, $DrugSim$\\
	\textbf{Input-2:} Disease-Disease similarity matrix, $DisSim$\\
	\textbf{Output:} Probability of treatment of disease $j$ by drug $i$, $ \hat{R}_{ij}$
	
	\begin{algorithmic}[1]
		
		\State Sample $N_p$ tuples from the set of validated drug-disease associations with label 1
		\State Sample another $N_p$ tuples from the set of validated drug-disease associations with label 0
		\State Sample $N_u$ tuples from the set of unvalidated drug-disease associations  with label 0
		\For {$(i,j)\in$ above sampled drug-disease associations}

		\State Extracting latent factor of drug and disease.
		\State \quad $  \big\{ d_{i}, d_{k} \} \leftarrow f(W \big\{ R_{i*}^T, R_{k*}^T \}+ b_1) $
		\State \quad $s_{j} \leftarrow  f(V R^T_{j*} + b_2)$
		
		\State Inferring interaction matrix based on Outer-product
		
		\State \quad $E^{ij} \leftarrow d_i \otimes s_j$
		
		\State Neighborhood Collaboration
		\State \quad $E^{kj} \leftarrow d_k \otimes s_j$
		\State \quad $E \leftarrow E^{ij} + \alpha E^{kj}$
		
		\State Inferring the predicted drug-disease associations.
		
		\State \quad $\hat{R}_{ij} \leftarrow F(P^T E Q +b)$

		\State \textbf{Return $\hat{R}_{ij}$}
		\EndFor
	\end{algorithmic} 
\end{algorithm}

\begin{figure*}[t]
	\centering
	\subfigure[Gottlieb Dataset]{
		\includegraphics[width=6cm,height=4cm]{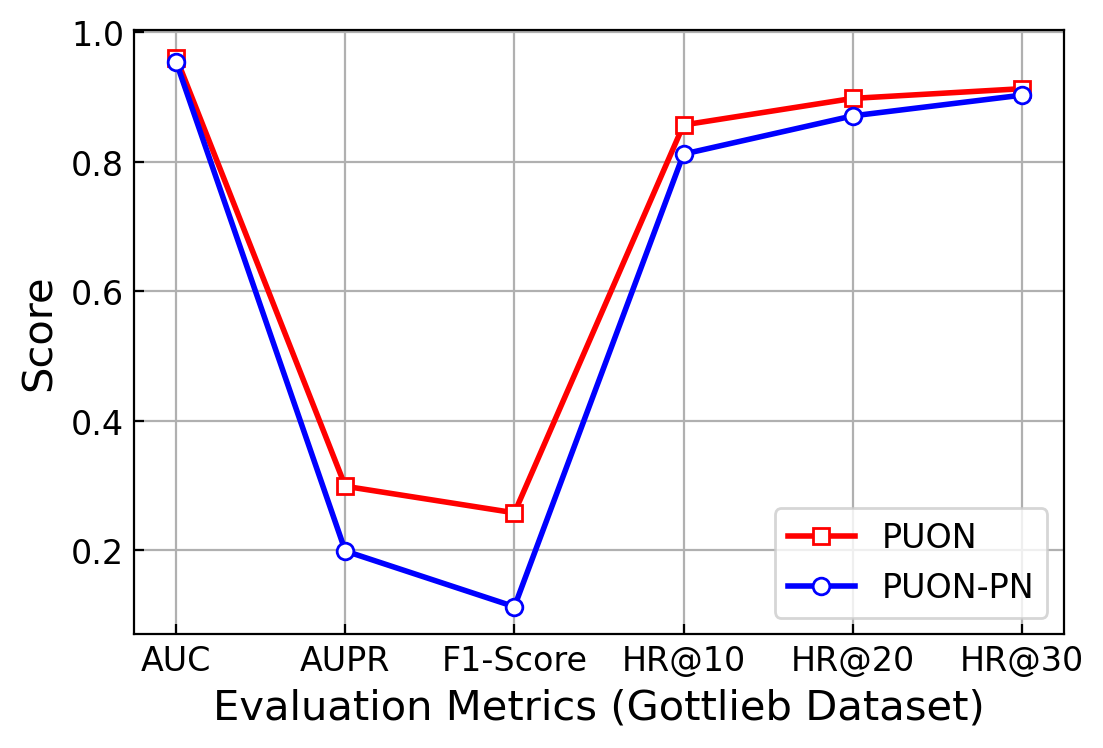}
	}
	\quad
	\subfigure[CDataset]{
		\includegraphics[width=6cm,height=4cm]{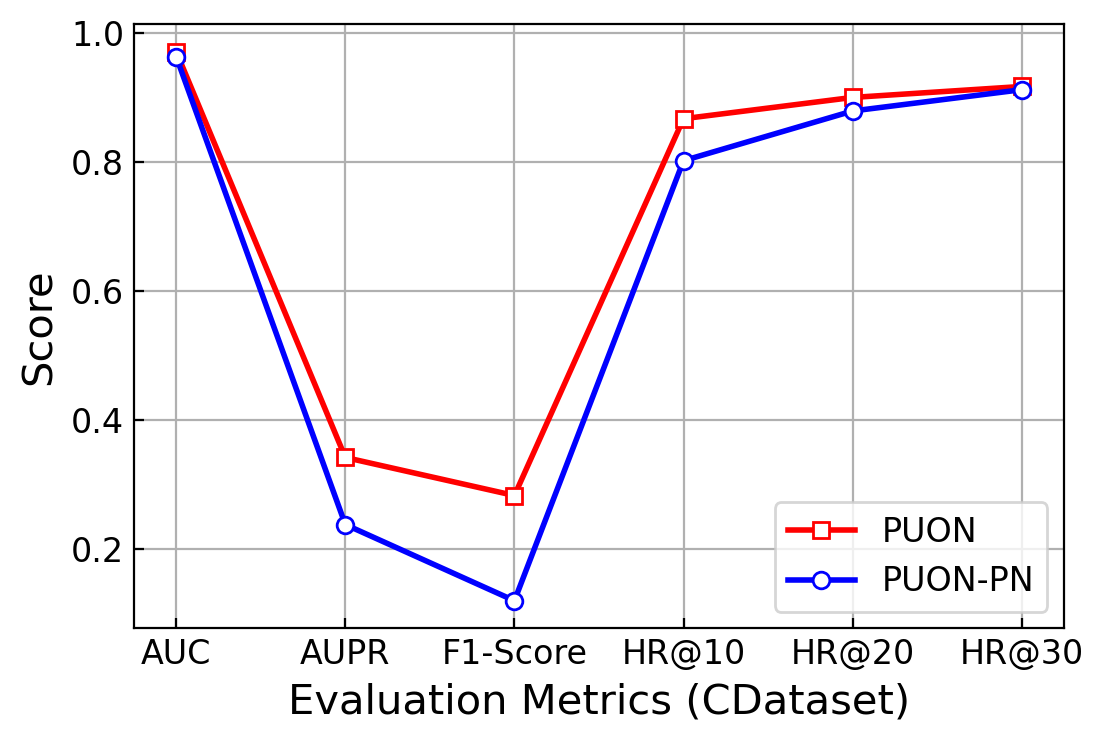}
	}

	\subfigure[DNDataset]{
		\includegraphics[width=6cm,height=4cm]{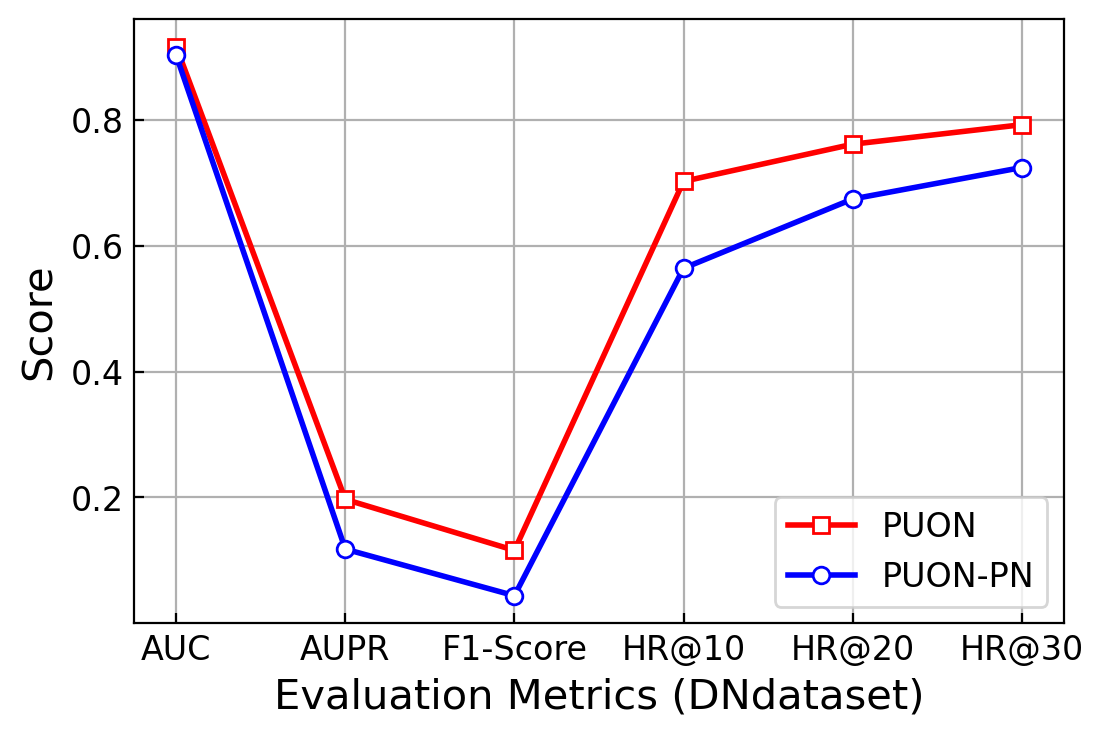}
	}
	\quad
	\subfigure[LDataset]{
		\includegraphics[width=6cm,height=4cm]{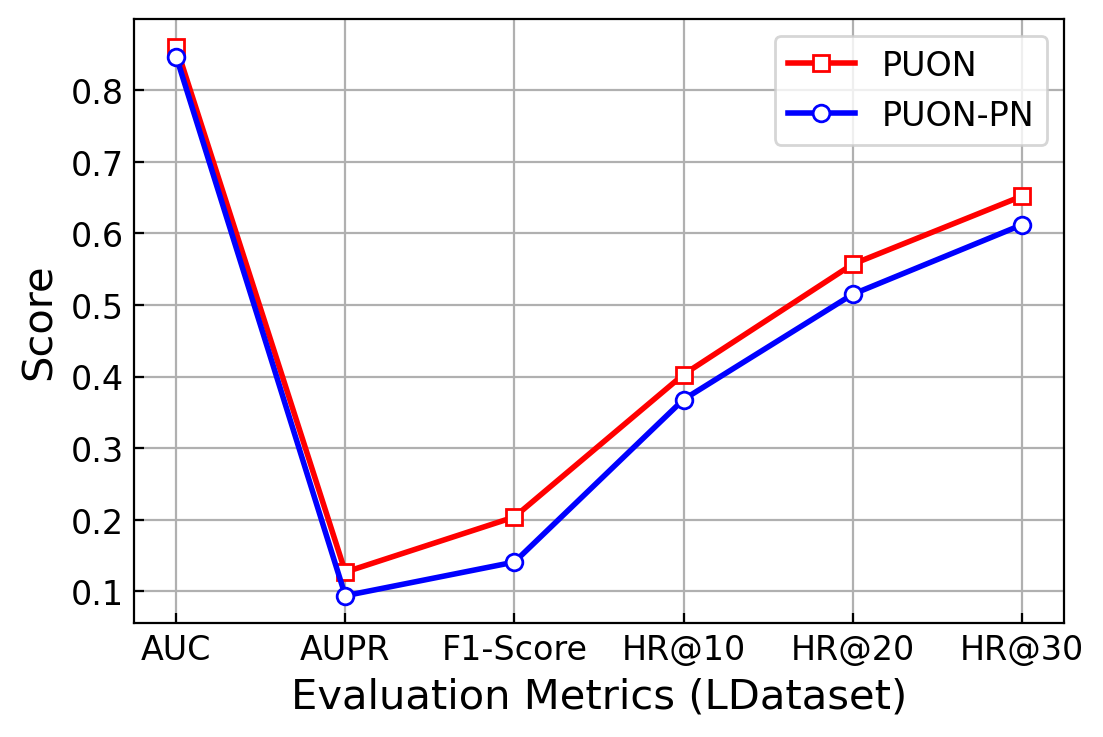}
	}
	\centering
	\caption{ The evaluation results of PUON and PUON-PN on four datasets}
\end{figure*}\par

\begin{figure*}[htbp]
	\centering
	\subfigure[Gottlieb Dataset]{
		\includegraphics[width=6cm,height=4cm]{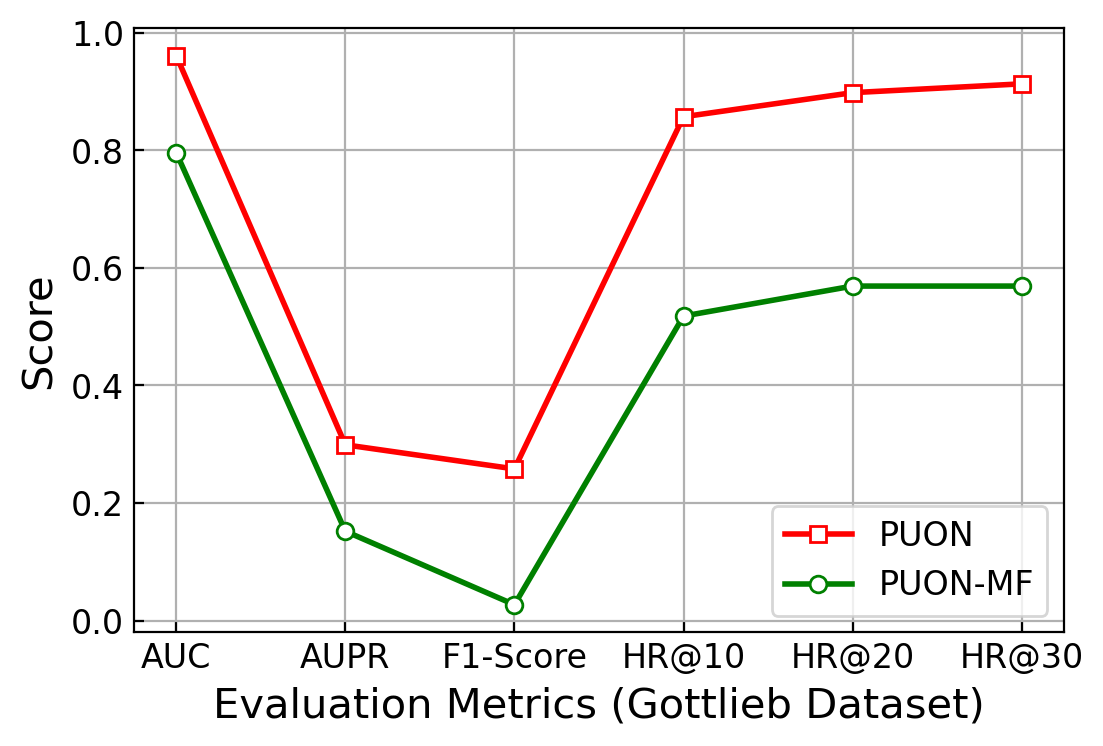}
	}
	\quad
	\subfigure[CDataset]{
		\includegraphics[width=6cm,height=4cm]{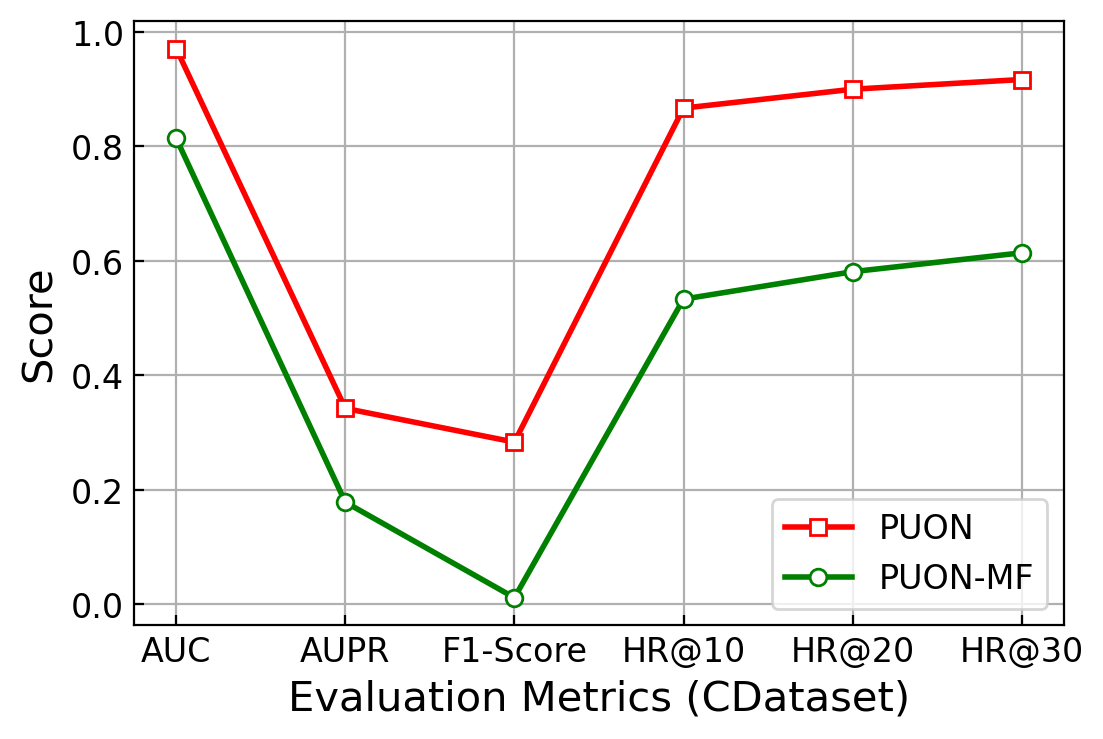}
	}
	
	\subfigure[DNDataset]{
		\includegraphics[width=6cm,height=4cm]{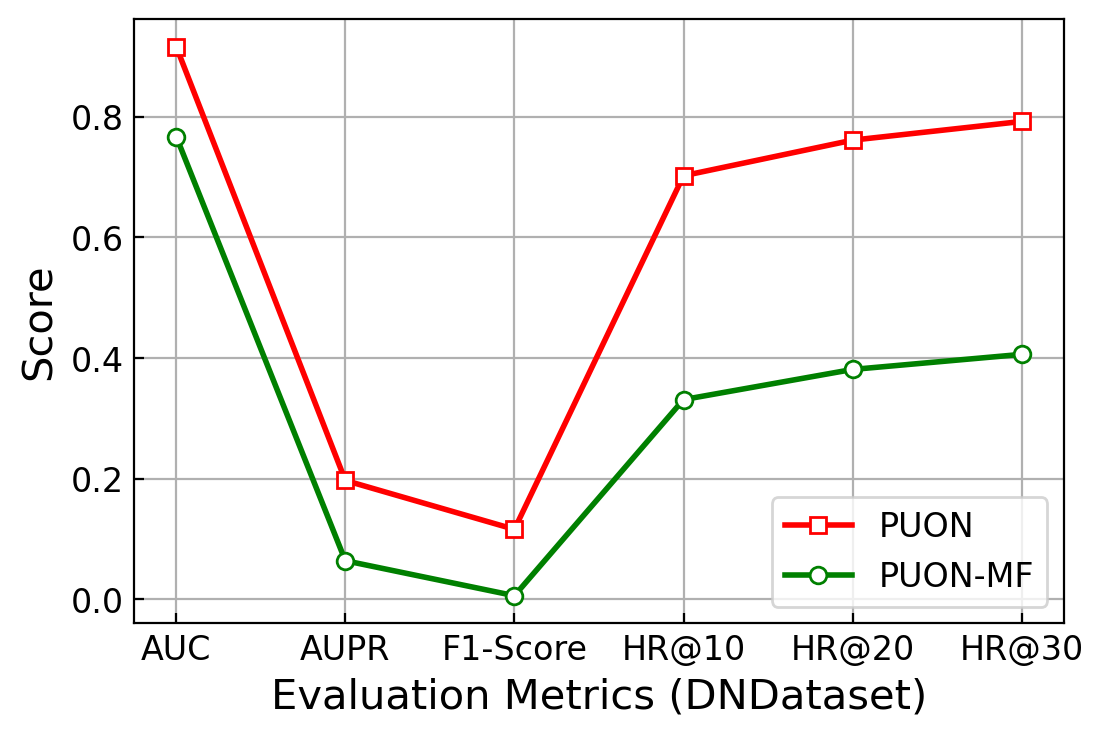}
	}
	\quad
	\subfigure[LDataset]{
		\includegraphics[width=6cm,height=4cm]{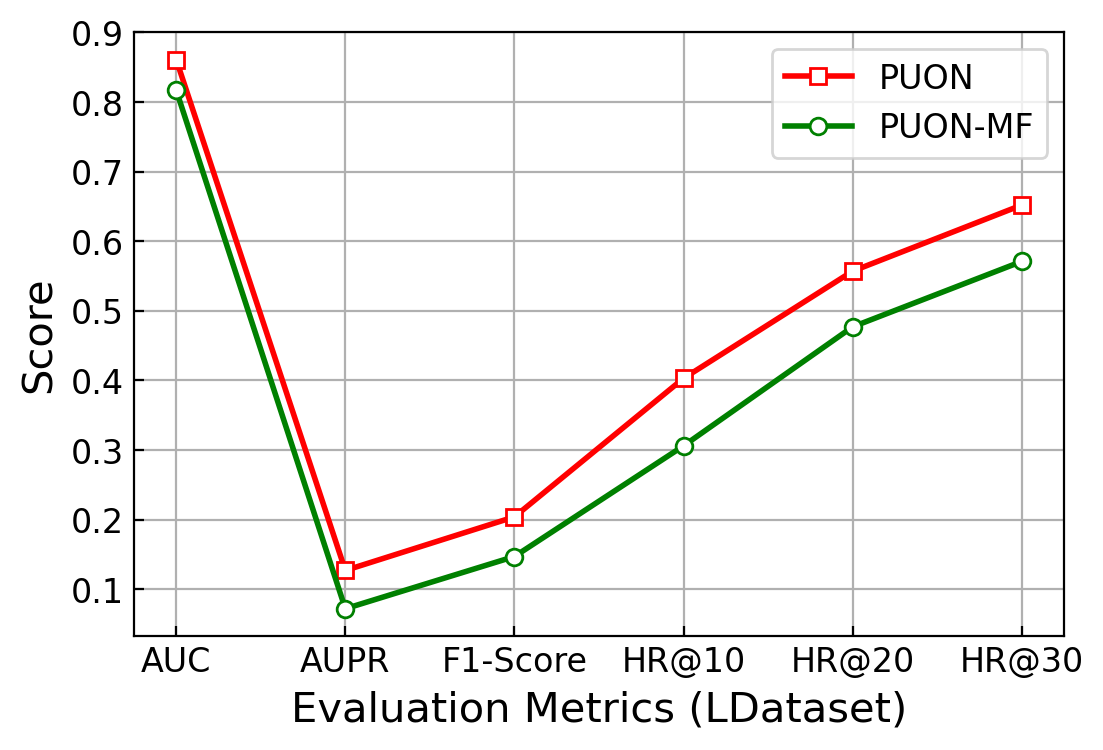}
	}
	\centering
	\caption{ The evaluation results of PUON and PUON-MF on four datasets}
\end{figure*}\par

\section{Experiments and Discussion}

We run several relevant experiments on four real-world drug-disease association datasets for evaluating the classification performance of PUON model. The experiments were aimed at answering the following four research questions: \par

\begin{itemize}
	\item[\textbf{RQ1}] Can the risk estimator without negative samples proposed in PUON help to improve the model performance?
	\item[\textbf{RQ2}] Does the proposed outer neighborhood-based classifier in PUON improve the performance of the model?
	\item[\textbf{RQ3}] Can the PUON model outperform state-of-the-art computational drug repositioning models?
	\item[\textbf{RQ4}] What can PUON offer in real-world scenarios to help drug development?
\end{itemize}

\begin{table}[h]
	\caption{The number of positive and negative samples used for training at each benchmark database }
	\centering
	\setlength{\tabcolsep}{7mm}
	\begin{tabular}{@{}c|cc@{}}
		\toprule
		Datasets  & \multicolumn{2}{c}{Training samples} \\ \midrule
		& Positive samples  & Negative samples \\
		Gottlieb  & 1740              & 19140            \\
		Cdataset  & 2279              & 25069            \\
		DNdataset & 908               & 9988             \\
		Ldataset  & 16575             & 182325           \\ \bottomrule
	\end{tabular}
\end{table}

\subsection{Evaluation Metrics}

For each of the above datasets, we use 10-fold cross-validation to assess the predictive effectiveness of the model. For all samples in the test set, feeding them into the PUON model yields a scalar value that is the predicted probability of the existence of a treatment relationship for the drug-disease association. So all samples in the test set are given a predicted value. The number of positive and negative samples used for training at each benchmark database is shown in Table 2. In addition, We performed 5 times of 10-fold cross-validation for all experiments, and took the average of 5 times as the final experimental result. \par

The experimental results of the model need to be compared numerically, so the evaluation metrics of AUC (Area Under Curve), AUPR (Area Under Precision-Recall Curve) and F1-Score are used in this work. These are the cornerstone metric currently applied to binary classification problems. In addition, we added hit ratio ($HR@K$) as the evaluation metric, where $K$ values are taken as 10, 20, and 30. In other words, three additional evaluation metrics are added as $HR@10$, $HR@20$, and $HR@30$. HR@K is calculated as shown below.\par

\begin{equation}
HR@K = \frac{\sum_{(i,j) \in p} hit(i)}{N}
\end{equation}

$p$ is the set of all validated drug-disease associations in the test set. For a validated drug-disease association $(i,j)$ in the test set, a list containing $k$ diseases is recommended for drug $i$. If disease $j$ exists in that list, $hit( i)$ is set to 1, otherwise $hit(i)$ is set to 0. $N$ represents the number of all validated drug-disease associations in the test set. \par

\subsection{Parameter Setting}

In this work, the range of variation of all hyperparameters in PUON is described below. The variation intervals of the dimensions of $W$ and $V$ in Equation (5) and Equation (6) are [8,16,32,64,128], whose values mainly control the expressiveness of the drug and disease latent factor. The variation interval of the learning rate of the optimizer is [0.001,0.01,0.1], which controls the convergence speed of PUON. The variation interval of the sampling number of unlabeled samples is [1,5,10,15,20], which controls the number of unlabeled data in the training set. In addition, PUON and other comparative model implementations are based on the Pytorch library. The default values of the above hyperparameters are 64, 0.001, 5, and 10. \par

\subsection{Effective of Risk Estimator in PUON (RQ1)}

To investigate whether the risk estimators without negative samples we propose in this work contribute to the improvement in model performance, we run relevant experiments comparing PUON with its variant version, PUON-PN. The difference between PUON and PUON-PN is that later uses negative sampling techniques to mark unvalidated drug-disease associations as negative samples, which may result in false samples. While PUON trains the parameters contained in the model using positive and unlabeled samples without negative sampling techniques. \par

The experimental results of PUON and PUON-PN on the four datasets are shown in Figure 3. Combining the performance under the six evaluation indicators, PUON has an average improvement of 31.5\%, 31.9\%, 46.4\% and 17.6\% compared with PUON-PN on the four datasets. It is worth noting that PUON has the largest improvement 106\% in F1-Score, which indicates that PUON is more suitable for drug repositioning scenarios. The above experimental results show that the risk estimator without negative samples fundamentally overcomes the risk generated by the negative sampling techniques (wrong samples) and avoids delaying the performance of the model. \par

Computational drug repositioning is a binary classification problem and suffers from category imbalance. Since the AUC metric is less sensitive to the class imbalance problem, it usually has a more optimistic estimate for the model. Therefore, the performance of PUON and PUON-PN under the AUC indicator is basically the same. However, AUPR and F1-Score can usually have a more objective evaluation of the prediction performance of the model in the class-imbalanced dataset, so the performance of PUON under AUPR and F1-Score is significantly better than that of PUON-PN. Therefore, the above phenomenon is not caused by the number of positive and negative samples. \par

 \par

\subsection{Effective of Classifier in PUON (RQ2)}

As mentioned above, inner product cannot take into account the cross information between each dimension of the latent factor (drug or disease). As a solution, the outer neighborhood-based classifier $C$ in PUON considers the cross information and neighborhood information between the dimensions of the latent facotr (drug or disease). Therefore, to investigate whether the outer neighborhood-based classifier $C$ in PUON can overcome the limitation of the inner product based model (matrix factorization), we compare PUON with the following variant model, PUON-MF. The difference between PUON and PUON-MF is that the latter uses inner product operation to generate the predicted values based on the latent factor of drug and disease. While the former uses outer neighborhood operation to generate the predicted values. And for a fair comparison, the same input is used for both. \par

The experimental results of PUON and PUON-MF on the four datasets are shown in Figure 4. we can find that the experimental results of the PUON model outperform the PUON-MF model on four data sets and with six evaluation metrics. The above experimental results illustrate that PUON can capture deep drug-disease associations more flexibly due to the inclusion of cross-information of the latent factor dimensions in the outer neighborhood-based classifier $C$. In addition, compared with Gottlieb dataset and LDataset, the PUON model has a larger improvement in CDataset and DNDataset with larger sparsity. This phenomenon implies that the introduction of neighborhood information overcomes the data sparsity problem. \par

\begin{table*}[t]
	\caption{Experimental results of the comparison models on four data sets}
	\centering
	\setlength{\tabcolsep}{2.7mm}
	\begin{tabular}{@{}ccccccccccc@{}}
		\toprule
		\multicolumn{1}{l}{}                           Dataset & Evaluation Metrics            & MF     & GMF    & SVM    & MLP    & ASMF   & NMFDR  & PUON-MF & PUON-PN & PUON   \\ \midrule
		\multicolumn{1}{c|}{\multirow{6}{*}{Gottlieb}}  & \multicolumn{1}{c|}{AUC}      & 0.705  & 0.817  & 0.724  & 0.822  & 0.891  & 0.887  & 0.796   & 0.954   & \textbf{0.961}  \\
		\multicolumn{1}{c|}{}                           & \multicolumn{1}{c|}{AUPR}     & 0.01   & 0.014  & 0.007  & 0.023  & 0.165  & 0.27   & 0.152   & 0.199   & \textbf{0.299}  \\
		\multicolumn{1}{c|}{}                           & \multicolumn{1}{c|}{F1-Score} & 0.018  & 0.036  & 0.021  & 0.042  & 0.221  & 0.232  & 0.027   & 0.113   & \textbf{0.258}  \\
		\multicolumn{1}{c|}{}                           & \multicolumn{1}{c|}{HR@10}    & 25.2\% & 34.3\% & 26.3\% & 35.5\% & 82.3\& & 83.2\% & 51.8\%  & 81.2\%  & \textbf{85.7\%} \\
		\multicolumn{1}{c|}{}                           & \multicolumn{1}{c|}{HR@20}    & 34.1\% & 45.2\% & 35.7\% & 46.1\% & 87.7\% & 87.9\% & 56.9\%  & 87.1\%  & \textbf{89.8\%} \\
		\multicolumn{1}{c|}{}                           & \multicolumn{1}{c|}{HR@30}    & 42\%   & 52.3\% & 42.6\% & 53.2\% & 89.2\% & 89.5\% & 60.2\%  & 90.3\%  & \textbf{91.3\%} \\ \midrule
		\multicolumn{1}{c|}{\multirow{6}{*}{Cdataset}}  & \multicolumn{1}{c|}{AUC}      & 0.757  & 0.868  & 0.763  & 0.872  & 0.911  & 0.927  & 0.814   & 0.963   & \textbf{0.971}  \\
		\multicolumn{1}{c|}{}                           & \multicolumn{1}{c|}{AUPR}     & 0.039  & 0.051  & 0.044  & 0.052  & 0.232  & 0.267  & 0.178   & 0.238   & \textbf{0.342}  \\
		\multicolumn{1}{c|}{}                           & \multicolumn{1}{c|}{F1-Score} & 0.025  & 0.047  & 0.027  & 0.038  & 0.264  & 0.278  & 0.011   & 0.12    & \textbf{0.283}  \\
		\multicolumn{1}{c|}{}                           & \multicolumn{1}{c|}{HR@10}    & 26.4\% & 38.9\% & 30.7\% & 41.7\% & 83.7\% & 84.9\% & 53.3\%  & 80.2\%  & \textbf{86.7\%} \\
		\multicolumn{1}{c|}{}                           & \multicolumn{1}{c|}{HR@20}    & 38.4\% & 49.3\% & 41.4\% & 50.2\% & 88.3\% & 88.8\% & 58.1\%  & 87.9\%  & \textbf{90\%}   \\
		\multicolumn{1}{c|}{}                           & \multicolumn{1}{c|}{HR@30}    & 44.6\% & 55.6\% & 45.9\% & 58.1\% & 89.7\% & 90.2\% & 61.4\%  & 91.2\%  & \textbf{91.7\%} \\ \midrule
		\multicolumn{1}{c|}{\multirow{6}{*}{DNdataset}} & \multicolumn{1}{c|}{AUC}      & 0.701  & 0.721  & 0.752  & 0.736  & 0.856  & 0.821  & 0.767   & 0.902   & \textbf{0.916}  \\
		\multicolumn{1}{c|}{}                           & \multicolumn{1}{c|}{AUPR}     & 0.009  & 0.01   & 0.007  & 0.011  & 0.051  & 0.057  & 0.064   & 0.118   & \textbf{0.197}  \\
		\multicolumn{1}{c|}{}                           & \multicolumn{1}{c|}{F1-Score} & 0.003  & 0.044  & 0.036  & 0.051  & 0.107  & 0.112  & 0.006   & 0.044   & \textbf{0.116}  \\
		\multicolumn{1}{c|}{}                           & \multicolumn{1}{c|}{HR@10}    & 20.8\% & 21.6\% & 20.9\% & 23.9\% & 60.9\% & 64.1\% & 33.1\%  & 56.4\%  & \textbf{70.2\%} \\
		\multicolumn{1}{c|}{}                           & \multicolumn{1}{c|}{HR@20}    & 25.5\% & 26.1\% & 26.5\% & 29.7\% & 68.4\% & 71.3\% & 38.1\%  & 67.4\%  & \textbf{76.1\%} \\
		\multicolumn{1}{c|}{}                           & \multicolumn{1}{c|}{HR@30}    & 28.6\% & 30.9\% & 31.8\% & 33.4\% & 72.9\% & 74.7\% & 40.6\%  & 72.4\%  & \textbf{79.2\%} \\ \midrule
		\multicolumn{1}{c|}{\multirow{6}{*}{Ldataset}}  & \multicolumn{1}{c|}{AUC}      & 0.763  & 0.801  & 0.778  & 0.799  & 0.827  & 0.812  & 0.817   & 0.847   & \textbf{0.861}  \\
		\multicolumn{1}{c|}{}                           & \multicolumn{1}{c|}{AUPR}     & 0.052  & 0.067  & 0.057  & 0.061  & 0.061  & 0.078  & 0.072   & 0.094   & \textbf{0.127}  \\
		\multicolumn{1}{c|}{}                           & \multicolumn{1}{c|}{F1-Score} & 0.101  & 0.125  & 0.117  & 0.116  & 0.013  & 0.027  & 0.147   & 0.141   & \textbf{0.204}  \\
		\multicolumn{1}{c|}{}                           & \multicolumn{1}{c|}{HR@10}    & 25.2\% & 28.7\% & 26.1\% & 27.5\% & 36.9\% & 38.1\% & 30.6\%  & 36.8\%  & \textbf{40.3\%} \\
		\multicolumn{1}{c|}{}                           & \multicolumn{1}{c|}{HR@20}    & 38.8\% & 42\%   & 39.8\% & 41.5\% & 52.3\% & 53.8\% & 47.7\%  & 51.5\%  & \textbf{55.7\%} \\
		\multicolumn{1}{c|}{}                           & \multicolumn{1}{c|}{HR@30}    & 48.4\% & 52.2\% & 49.9\% & 51.4\% & 62.1\% & 63.3\% & 57.1\%  & 61.2\%  & \textbf{65.2\%} \\ \bottomrule
	\end{tabular}
\end{table*}

\subsection{Performance Comparison RQ(3)}

We compare PUON with the following 8 popular models. \par

\begin{table*}[t]
	\caption{Experimental results of the comparison models on four data sets (New drug scenario)}
	\centering
	\setlength{\tabcolsep}{2.7mm}
	\begin{tabular}{@{}ccccccccccc@{}}
		\toprule
		Dataset                                         & Evaluation Metrics            & MF     & GMF    & SVM    & MLP    & ASMF   & NMFDR  & PUON-MF & PUON-PN & PUON   \\ \midrule
		\multicolumn{1}{c|}{\multirow{6}{*}{Gottlieb}}  & \multicolumn{1}{c|}{AUC}      & 0.791  & 0.854  & 0.812  & 0.852  & 0.917  & 0.876  & 0.892   & 0.935   & \textbf{0.942}  \\
		\multicolumn{1}{c|}{}                           & \multicolumn{1}{c|}{AUPR}     & 0.004  & 0.009  & 0.005  & 0.007  & 0.018  & 0.021  & 0.018   & 0.037   & \textbf{0.103}  \\
		\multicolumn{1}{c|}{}                           & \multicolumn{1}{c|}{F1-Score} & 0.004  & 0.027  & 0.007  & 0.026  & 0.029  & 0.037  & 0.031   & 0.058   & \textbf{0.063}  \\
		\multicolumn{1}{c|}{}                           & \multicolumn{1}{c|}{HR@10}    & 15.7\% & 23.3\% & 16.9\% & 21\%   & 43.8\% & 73\%   & 17.5\%  & 75.4\%  & \textbf{77.1\%} \\
		\multicolumn{1}{c|}{}                           & \multicolumn{1}{c|}{HR@20}    & 21.3\% & 27.4\% & 22.7\% & 25.1\% & 54.3\% & 79.2\% & 28.6\%  & 78.3\%  & \textbf{81.8\%} \\
		\multicolumn{1}{c|}{}                           & \multicolumn{1}{c|}{HR@30}    & 27.6\% & 33.9\% & 29.2\% & 32.1\% & 60.2\% & 80.7\% & 36.8\%  & 80.1\%  & \textbf{82.4\%} \\ \midrule
		\multicolumn{1}{c|}{\multirow{6}{*}{Cdataset}}  & \multicolumn{1}{c|}{AUC}      & 0.787  & 0.845  & 0.801  & 0.841  & 0.911  & 0.862  & 0.892   & 0.93    & \textbf{0.935}  \\
		\multicolumn{1}{c|}{}                           & \multicolumn{1}{c|}{AUPR}     & 0.003  & 0.015  & 0.003  & 0.007  & 0.007  & 0.006  & 0.005   & 0.02    & \textbf{0.023}  \\
		\multicolumn{1}{c|}{}                           & \multicolumn{1}{c|}{F1-Score} & 0.005  & 0.011  & 0.007  & 0.011  & 0.045  & 0.021  & 0.015   & 0.038   & \textbf{0.048}  \\
		\multicolumn{1}{c|}{}                           & \multicolumn{1}{c|}{HR@10}    & 16.3\% & 19.5\% & 17.8\% & 18.6\% & 56.4\% & 61.6\% & 29.9\%  & 71.1\%  & \textbf{75.7\%} \\
		\multicolumn{1}{c|}{}                           & \multicolumn{1}{c|}{HR@20}    & 23.7\% & 26.2\% & 25.2\% & 25.4\% & 61.5\% & 64.4\% & 38.4\%  & 74\%    & \textbf{80.2\%} \\
		\multicolumn{1}{c|}{}                           & \multicolumn{1}{c|}{HR@30}    & 25.4\% & 31\%   & 26.9\% & 27.6\% & 66.1\% & 68.9\% & 40.1\%  & 76.8\%  & \textbf{81.3\%} \\ \midrule
		\multicolumn{1}{c|}{\multirow{6}{*}{DNdataset}} & \multicolumn{1}{c|}{AUC}      & 0.67   & 0.724  & 0.696  & 0.716  & 0.759  & 0.726  & 0.741   & 0.765   & \textbf{0.772}  \\
		\multicolumn{1}{c|}{}                           & \multicolumn{1}{c|}{AUPR}     & 0.005  & 0.008  & 0.004  & 0.006  & 0.012  & 0.009  & 0.007   & 0.014   & \textbf{0.021}  \\
		\multicolumn{1}{c|}{}                           & \multicolumn{1}{c|}{F1-Score} & 0.007  & 0.035  & 0.009  & 0.028  & 0.037  & 0.02   & 0.03    & 0.073   & \textbf{0.087}  \\
		\multicolumn{1}{c|}{}                           & \multicolumn{1}{c|}{HR@10}    & 10.3\% & 17.7\% & 12.5\% & 19.8\% & 44.1\% & 49.5\% & 20.9\%  & 51.5\%  & \textbf{53.6\%} \\
		\multicolumn{1}{c|}{}                           & \multicolumn{1}{c|}{HR@20}    & 16.1\% & 23.1\% & 17.2\% & 21.3\% & 51.9\% & 55.9\% & 27.8\%  & 57.3\%  & \textbf{58.5\%} \\
		\multicolumn{1}{c|}{}                           & \multicolumn{1}{c|}{HR@30}    & 21.3\% & 27\%   & 23.4\% & 25.9\% & 54.1\% & 57.3\% & 31.7\%  & 59.9\%  & \textbf{61\%}   \\ \midrule
		\multicolumn{1}{c|}{\multirow{6}{*}{Ldataset}}  & \multicolumn{1}{c|}{AUC}      & 0.803  & 0.927  & 0.828  & 0.939  & 0.835  & 0.847  & 0.827   & 0.941   & \textbf{0.949}  \\
		\multicolumn{1}{c|}{}                           & \multicolumn{1}{c|}{AUPR}     & 0.024  & 0.101  & 0.036  & 0.105  & 0.069  & 0.052  & 0.072   & 0.181   & \textbf{0.209}  \\
		\multicolumn{1}{c|}{}                           & \multicolumn{1}{c|}{F1-Score} & 0.06   & 0.076  & 0.071  & 0.084  & 0.074  & 0.067  & 0.066   & 0.092   & \textbf{0.109}  \\
		\multicolumn{1}{c|}{}                           & \multicolumn{1}{c|}{HR@10}    & 10.9\% & 18\%   & 12.1\% & 17.6\% & 55\%   & 50.5\% & 31.7\%  & 53.8\%  & \textbf{57.4\%} \\
		\multicolumn{1}{c|}{}                           & \multicolumn{1}{c|}{HR@20}    & 14.7\% & 28.9\% & 16.7\% & 28.1\% & 71.9\% & 67.9\% & 40.3\%  & 71.6\%  & \textbf{74.5\%} \\
		\multicolumn{1}{c|}{}                           & \multicolumn{1}{c|}{HR@30}    & 22.9\% & 35.8\% & 25.7\% & 35.6\% & 73.9\% & 74.7\% & 41.1\%  & 80.1\%  & \textbf{81.6\%} \\ \bottomrule
	\end{tabular}
\end{table*}

\begin{itemize}
	
	\item \emph{MF} \cite{mf}: The matrix factorization model uses two latent factor matrixes for representing drugs and diseases, respectively, and subsequently performs an inner product operation on the two matrixes to derive the predicted drug-disease association.
	\item \emph{GMF} \cite{xn3}: The generalized matrix factorization model is similar to the MF, with the difference that it improves the inner product operation and adaptively adjusts the weights of each dimension of the latent factor through the neural network to have better prediction ability.
	
	\item \emph{SVM} \cite{svm}: The support vector machine is a classic machine learning classification model, which is not disturbed by outliers and has good generalization ability.
	
	\item \emph{MLP} \cite{mlp}: The multilayer perceptrons are composed of multiple neural networks, which can extract nonlinear characteristics of drugs and diseases, so that new effects of drugs can be predicted.

	
		

	\item \emph{ASMF} \cite{asmf}: ASMF designed a new self-encoder model, which is able to introduce drug-disease association, drug similarity information and disease similarity information into the latent factor extraction process. The latent factor obtained above has a good expression capability, which allows the calculation of prediction probabilities with high generalizability.
	
	\item \emph{NMFDR} \cite{nmfdr}: The NMFDR model uses the generalized Euclidean distance to model the association between drugs and diseases, and also uses metric information to limit the feature space of drugs and diseases to obtain a high-quality feature vector.

	\item \emph{PUON-PN}: The PUON-PN model is a variant of our proposed PUON, with the difference that the PUON-PN model utilizes negative sampling techniques for data sampling.
	
	\item \emph{PUON-MF}: The PUON-MF model is based on the PUON framework, with the difference that PUON-MF utilizes a matrix factorization model to replace the classifier $C$ in PUON.
	
\end{itemize}

We can draw the following conclusions from the results in Table 3. \par

The matrix factorization model achieved the worst experimental results under multiple metrics on both datasets, indicating that the inner product operation is unable to capture complex drug-disease associations. The GMF model has better experimental results than the matrix factorization model with fixed weights due to the introduction of adjustable weight parameters. This indicates that the prediction performance of the model can be improved by changing the weight values of different dimensions. \par

The prediction effect of the MLP model is better than that of the SVM model, which indicates that the neural network-based model is capable of mining complex drug-disease associations and is more suitable for drug repositioning scenarios. In addition, ASMF and NMFDR are the current SOTA models, the experimental results of both ASMF and NMFDR models outperformed the other models and were second only to the experimental results of our proposed PUON model. This also indicates the SOTA level of these two models. \par

Finally, it is worth noting that our proposed PUON model achieves first place under several metrics, which highlights the advantages of our proposed risk estimator applicable to the computational drug repositioning domain and the classifier using outer collaboration. The PUON model outperforms PUON-PN model, indicating that the negative sampling technique may produce erroneous samples and thus reduce the performance of the model. The PUON model without negative sampling technique, on the other hand, does not produce erroneous samples, thus improving the predictive power of the model, which highlights the superiority of the risk estimator in the PUON over the negative sampling technique. And the prediction ability of the classifier $C$ based on the outer collaboration is better than the matrix factorization model and the GMF model, which indicates that the outer product has more degrees of freedom and crossover information. Thus the above experimental results highlight the effectiveness of the improvement points in this work. \par

\begin{table*}[]
	\caption{The top 5 candidate indications for 5 specific drugs}
	\centering
	\setlength{\tabcolsep}{6mm}
	\begin{tabular}{ll}
		\hline
		Drugs (DrugBank IDs)                                       & Top 5 candidate diseases (OMIM IDs)                                        \\ \hline
		\multicolumn{1}{c}{\multirow{5}{*}{Doxorubicin (DB00997)}} & \textbf{Kaposi's sarcoma, susceptibility (148000)};                                \\
		\multicolumn{1}{c}{}                                       & Spastic paraplegia and EVANS syndrome (601608);                             \\
		\multicolumn{1}{c}{}                                       & Testicular germ cell tumor (273300);                                        \\
		\multicolumn{1}{c}{}                                       & Reticular cell sarcoma (267730);                                            \\
		\multicolumn{1}{c}{}                                       & \textbf{Esophageal cancer (133239)};                                                 \\ \hline
		\multirow{5}{*}{Gemcitabine (DB00441)}                     & Mismatch Repair Cancer Syndrome 1 (276300);                                 \\
		& \textbf{Lymphoma, Hodgkin, classic (236000)};                                    \\
		& Leukemia, chronic lymphocytes, susceptibility, 2 (109543);                  \\
		& Reticular cell sarcoma (267730);                                            \\
		& HAJDU-CHENEY syndrome (102500);                                             \\ \hline
		\multirow{5}{*}{Vincristine (DB00541)}                     & Developmental delay, small and swollen hands and feet, and eczema (233810); \\
		& \textbf{Kaposi Sarcoma, Susceptibility to (148000)};                             \\
		& \textbf{Small Cell Cancer of The Lung (182280)};                              \\
		& Testicular germ cell tumor (273300);                                        \\
		& Mycosis Fungoides (254400);                                                 \\ \hline
		\multirow{5}{*}{Methotrexate (DB00563)}                    & \textbf{Multiple myeloma (254500)};                                              \\
		& Neuroblastoma, susceptibility, 1 (256700);                                  \\
		& Thrombocytopenic purpura, autoimmune (188030);                              \\
		& Renal cell carcinoma, non-capillary (144700);                               \\
		& Kaposi Sarcoma, Susceptibility to (148000);                                 \\ \hline
		\multirow{5}{*}{Risperidone (DB00734)}                     & \textbf{Obsessive-Compulsive Disorder (164230)};                                 \\
		& Alcohol dependence (103780);                                                \\
		& Camurati-Engelman disease (131300);                                         \\
		& Mitral valve prolapse 1 (157700);                                           \\
		& \textbf{Panic disorder 1 (167870)};                                              \\ \hline
	\end{tabular}
\end{table*}

\subsection{Predicting indication for new drug}

The new drug scenario is that the drug does not have any known therapeutic disease. We construct the new drug scenario in the following way, there are 171 drugs in the Gottlieb dataset with only one known therapeutic disease, and we use these 171 drug-disease associations as the test set and the rest of the drug-disease associations as the training set. Similarly, the same process is performed for Cdataset, DNdataset and Ldataset. \par

Table 4 shows the experimental results of all the comparison methods in the new drug scenario. We can find that the PUON model has achieved the best prediction results on the four real-world datasets. Firstly, compared with the PUON-PN model, the average improvement of PUON on the four datasets is 32.8\%, 10.4\%, 13\% and 7.9\%, respectively. The average improvement value is obtained by accumulating the improvement values of all six evaluation metrics and averaging them. The above experimental results show that the risk estimator without the negative sampling technique has better generalization performance. \par

Secondly, the experimental results of the PUON model outperform the PUON-MF model on four data sets and with six evaluation metrics. Such experimental results show that compared with the inner product, the outer product and the neighborhood collaboration enable PUON to consider more feature information, resulting in a better prediction effect. Finally, the performance of PUON is better than other comparison algorithms, which verifies its superiority. \par

The advantage of PUON is that firstly, the model is trained using unlabeled samples instead of negative sampling technique, which avoids the generation of incorrect training samples. This is because the negative sampling technique tends to treat potentially positive samples as negative samples when sampling, resulting in the model learning the wrong information and thus causing degradation in prediction performance. Secondly, the use of outer neighborhood allows the model to take into account more feature information. \par

However, due to the imbalance of positive and negative sample categories in the computational drug repositioning dataset, the performance of PUON under the AUPR and F1-Score metrics is not good. This is also a research focus of our future work, that is, how to improve the prediction performance of the model in the case of category imbalance. \par

\subsection{Case Study (RQ4)}

In order to highlight the value of the PUON model for practical drug development, we conducted a case study of the marketed drugs in the Gottlieb dataset. Similar to the study by Yang et al\cite{m2}, the four antineoplastic drugs, doxorubicin, gemcitabine, vincristine and methotrexate, and the antipsychotic drug, risperidone, were selected. In the case study, we used all the validated drug-disease associations in the Gottlieb dataset for training the PUON model, and subsequently recommended 5 potential diseases from high to low for the above five drugs based on the model scoring. Finally, the CTD dataset was queried to verify whether there was a therapeutic relationship between the target drug and the potential diseases. \par

Table 5 lists the five drugs mentioned above and the set of potential diseases inferred from the PUON model that correspond to them. As we can see from Table 5, the recommended list for doxorubicin includes 2 diseases, Kaposi Sarcoma, Susceptibility to (148000), and Esophageal cancer (133239), which were verified in the CTD database. The list of recommended diseases for Gemcitabine and Methotrexate was validated in the CTD database for 1 disease each, Lymphoma, Hodgkin, classic (236000), and Multiple myeloma (254500), and both were predicted in the top order. For both Vincristine and Risperidone, two diseases were validated in the respective recommended disease lists. The above prediction set shows that PUON can predict right potential diseases, and most of the diseases that can be treated are ranked in the top order, which strongly indicates the value of PUON model in practical drug development. \par

The following content discusses the future improvement direction and application scenarios of the PUON model. \par

Drug combination is a popular and effective treatment for drug-resistant diseases. Chen et al. \cite{add1} proposed a novel synergistic drug combination prediction method. This method found 7 groups of synergistic drug combinations, which can effectively overcome fungal drug resistance. The PUON model of this work is essentially a binary classification model, i.e., given the features of the target drug and the target disease, and inputting them into PUON, the probability that the target drug can treat the target disease can be obtained, with 1 means that the drug can treat the disease and 0 means the drug cannot treat the disease. If we consider pairwise drug combinations, we can synthesize the features of two drugs and input them into the PUON model together with the features of the target disease, and we can obtain the probability that the drug combination can treat the disease. However, if the number of drugs is too large, a large number of drug combinations will be formed, which requires a great amount of computation. Therefore, in our future work, we will investigate how to synthesize the features of the drug combination and simplify the computational effort so that the PUON model can be appropriately applied to synergistic drug combination prediction. \par

The PUON model proposed in this work only uses known drug-disease associations in the process of mining the latent factor of drug. Learning a good latent factor of drug is the key to improving the predictive ability of the model. Therefore, in future work, we can integrate multi-source information such as drug-disease association network, drug-target association network and drug-microRNA association network into a new heterogeneous network, and then use algorithms such as random walk to mine the latent factor of drug. Through the above steps, a high-quality latent factor of drug can be learned, which contains more structural information. Chen et al. summarize the advantages and disadvantages of popular models for drug-target association prediction \cite{add2} and drug-microRNA association prediction \cite{add3}. And they \cite{add4} proposed a prediction model of small molecule drug-microRNA association based on bounded nuclear norm regularization. In our future work, we will use the above literature as a starting point to understand the latest research progress in drug-target association prediction and drug-microRNA association prediction, so that we can better combine them with drug-disease associations.

\section{Conclusion}

In this paper, we novelly propose PUON framework for computational drug repositioning to predict potential drug-disease associations without negative sampling technique. We also propose a outer neighborhood-based classifier that models the correlation-information between dimensions of latent factor. Then we analyze the effectiveness of the above improvement points through extensive experiments and validate the superiority of PUON in four real-world datasets by comparing it with 8 popular baselines under 6 evaluation metrics. In future work, we will explore how to introduce multi-source data of drugs and diseases for enhancing the generalization performance of the PUON.\par

\bibliographystyle{IEEEtran}
\bibliography{reference.bib}

\end{document}